# Designing a Dataset for Convolutional Neural Networks to Predict Space Groups Consistent with Extinction Laws

**Hao Wang [a, b], Jiajun Zhong [a, b], Yikun Li [a, b], Junrong Zhang [a, b, c], Rong Du [*, a, b]**

**a Institute of High Energy Physics, Chinese Academy of Sciences, Beijing 100049, China**

**b Spallation Neutron Source Science Center, Dongguan 523803, China**

**c University of Chinese Academy of Sciences, Beijing 100049, China**

***: Correspondence to: Rong Du, E-mail: durong@ihep.ac.cn**

**Abstract:**

In this paper, a dataset of one-dimensional powder diffraction patterns was designed with new strategy to train Convolutional Neural Networks for predicting space groups. The diffraction pattern was calculated based on lattice parameters and Extinction Laws, instead of the traditional approach of generating it from a crystallographic database. This paper demonstrates that the new strategy is more effective than the conventional method. As a result, the model trained on the cubic and tetragonal training set from the newly designed dataset achieves prediction accuracy that matches the theoretical maximums calculated based on Extinction Laws. These results demonstrate that machine learning-based prediction can be both physically reasonable and reliable. Additionally, the model trained on our newly designed dataset shows excellent generalization capability, much better than the one trained on a traditionally designed dataset.

# 1. Introduction

X-ray and neutron powder diffraction are common methods in materials science for exploring the crystal structures of materials. The powder diffraction data reflect a one-dimensional projection of the three-dimensional structural information (De Wolff, 1957; Mighell & Santoro, 1975). A critical aspect of analyzing powder diffraction data is the identification of space groups, which reflect the symmetry of the crystal structure. Given the finite number (230) of space groups, this task can be approached as a classification problem. The complexity of space group identification, combined with the robust classification capabilities of Machine Learning (ML) (Jordan & Mitchell, 2015), has made ML a promising tool of space group prediction. Numerous studies have demonstrated its potential (Park *et al.*, 2017; Garcia-Cardona *et al.*, 2019; Aguiar *et al.*, 2020; Liu *et al.*, 2019; Lee *et al.*, 2022; Suzuki *et al.*, 2020; Vecsei *et al.*, 2019; Chitturi *et al.*, 2021; Lolla *et al.*, 2022). The earliest research using ML to predict space groups from powder diffraction patterns was published in 2017. In this study, (Park *et al.*, 2017) attempted to use Convolutional Neural Networks (CNNs) to predict the crystal system, extinction group, and space group. They generated powder diffraction patterns from numerous crystal structures in the Inorganic Crystal Structure Database (ICSD) (Hellenbrandt, 2004) to train their model. The model achieved accuracies of 81.14%, 83.83%, and 94.99% for predicting the space group, extinction group, and crystal system, respectively. Following this, several researchers conducted studies on space group prediction using ML. (Vecsei *et al.*, 2019) compared the performance of CNNs and a simple dense network. They found that when both models were trained on ICSD-generated data, the simple dense network outperformed CNNs in predicting the space group of real data. (Lolla *et al.*, 2022) employed semi-supervised learning by integrating unlabeled experimental diffraction data into the training set, expanding the dataset and improving the model's prediction accuracy. Due to the lack of well-labeled experimental data, it is necessary to rely on extensive simulated data to train ML models. The simulated data is derived from numerous crystal structures available in several crystallographic databases, including the Inorganic Crystal Structure Database (ICSD) (Hellenbrandt, 2004), the Cambridge Structural Database (CSD) (Groom *et al.*, 2016), the Crystallography Open Database (COD) (Downs & Hall-Wallace, 2003; Gražulis *et al.*, 2009, 2012, 2015; Merkys *et al.*, 2016, 2023; Quirós *et al.*, 2018; Vaitkus *et al.*, 2021) and the Materials Project (MP) (Jain *et al.*, 2013). Recently, a research (Schopmans *et al.*, 2023) attempted to generate simulated data independently of existing databases and demonstrated that synthetically generated crystals also can be used to extract structural information from databases' powder diffraction patterns. Despite the focus of current studies on improving prediction accuracy, none has provided explanations for the accuracy of space group predictions, resulting in limited confidence in the results.

In this paper, we focus on dataset design rather than model architecture. We developed three distinct types of datasets. The first, termed the Crystallographic Database Derived Dataset (CDDD), is the conventional dataset used in space group prediction. The second, the Uniform Lattice Based Dataset (ULBD), is designed without relying on actual crystal structures. The third, the Generated Crystal Derived Dataset (GCDD), employs virtual crystal structures to simulate diffraction patterns. Our findings indicate that the CDDD is not ideal for training space group prediction models due to inherent feature biases and imbalances among space groups. Meanwhile, using the ULBD yields prediction accuracies consistent with extinction rules. This remarkable consistency strongly suggests that the reliability of models trained on the ULBD.

## 2. The used CNNs

A traditional CNN, closely resembling the model in (Lee *et al.*, 2022), was adopted. It includes multiple overlapping one-dimensional convolutional layers, max pooling, and several dropout layers. All CNNs in this study are based on this architecture, with output layer variations tailored to specific tasks.

## 3. The dataset

### 3.1 The Crystallographic Database Derived Dataset (CDDD)

The CDDD was generated using crystal structures obtained from the Materials Project. The generation process is illustrated in **Fig. 1(a)**. A total of 155,360 crystal structures were downloaded from the Materials Project. Then, the diffraction patterns of all structures were then calculated to construct the dataset. The peak positions and corresponding structure factors were initially obtained using Computational Crystallography Toolbox (cctbx) (Grosse-Kunstleve & Adams, 2003). The Lorentz correction was then applied to the structure factors. The corrected structure factors were treated as the peak intensities. Peak intensities and positions were used to construct line patterns. Finally, convolution between a peak shape function and the generated line patterns was performed to obtain the diffraction patterns.

## 3.2 The Uniformed Lattice Based Dataset (ULBD)

We propose a specially designed dataset called the Uniform Lattice Based Dataset (ULBD). The generation process is shown in **Fig. 1(b)**. Firstly, the range and step size for each lattice parameter are set for each space group. This results in a mesh grid for each space group, where each point in the mesh grid represents a group of lattice parameters (a, b, c, α, β, γ). Secondly, for each group of lattice parameters, the positions of the diffraction peaks are calculated using **Equation 1** (Mighell & Santoro, 1975). Thirdly, peaks without intensity are eliminated by considering the extinction rules specific to each space group. Fourthly, multiple sets of peak intensities are generated for each group of peak positions, and Lorentz correction is applied, followed by normalization of the maximum peak intensity.

$$Q_{hkl} = h^2a^{*2} + k^2b^{*2} + l^2c^{*2} + 2klb^*c^*cos\alpha^* + 2hla^*c^*cos\beta^* + 2hka^*b^*cos\gamma^* \quad (1)$$

As a result, we get multiple line patterns for each group of lattice parameters. Finally, convolution between a peak shape function and the generated line patterns is performed to produce the diffraction pattern. The key feature of ULBD is the uniform distribution of lattice parameters within each space group. Additionally, because the crystal structures are not used for generating the ULBD, the architecture of the dataset is more flexible, and the number of samples is not constrained. In the following sections, we will demonstrate that the ULBD is more effective than CDDD for training CNN models to predict space groups.

## 3.3 The Generated Crystal Derived Dataset (GCDD)

The GCDD is designed using the method similar to (Schopmans *et al.*, 2023), while also trying to keep the uniform distribution in the scale of space groups. Virtual crystal structures were generated using PyXtal (Fredericks *et al.*, 2021), as shown in **Fig. 1(c).** A unique chemical compound was randomly selected for each generation without reuse. Then, PyXtal generated a crystal structure based on the specified compound and space group. Subsequently, the space group analyzer from pymatgen (Ong *et al.*, 2013) was used to verify whether the space group of the generated structure matched the one initially specified. This process was repeated up to 1000 times for each space group or until PyXtal could no longer generate any new structures. The GCDD was generated based on these virtual structures with the same method for generating the CDDD and ULBD.

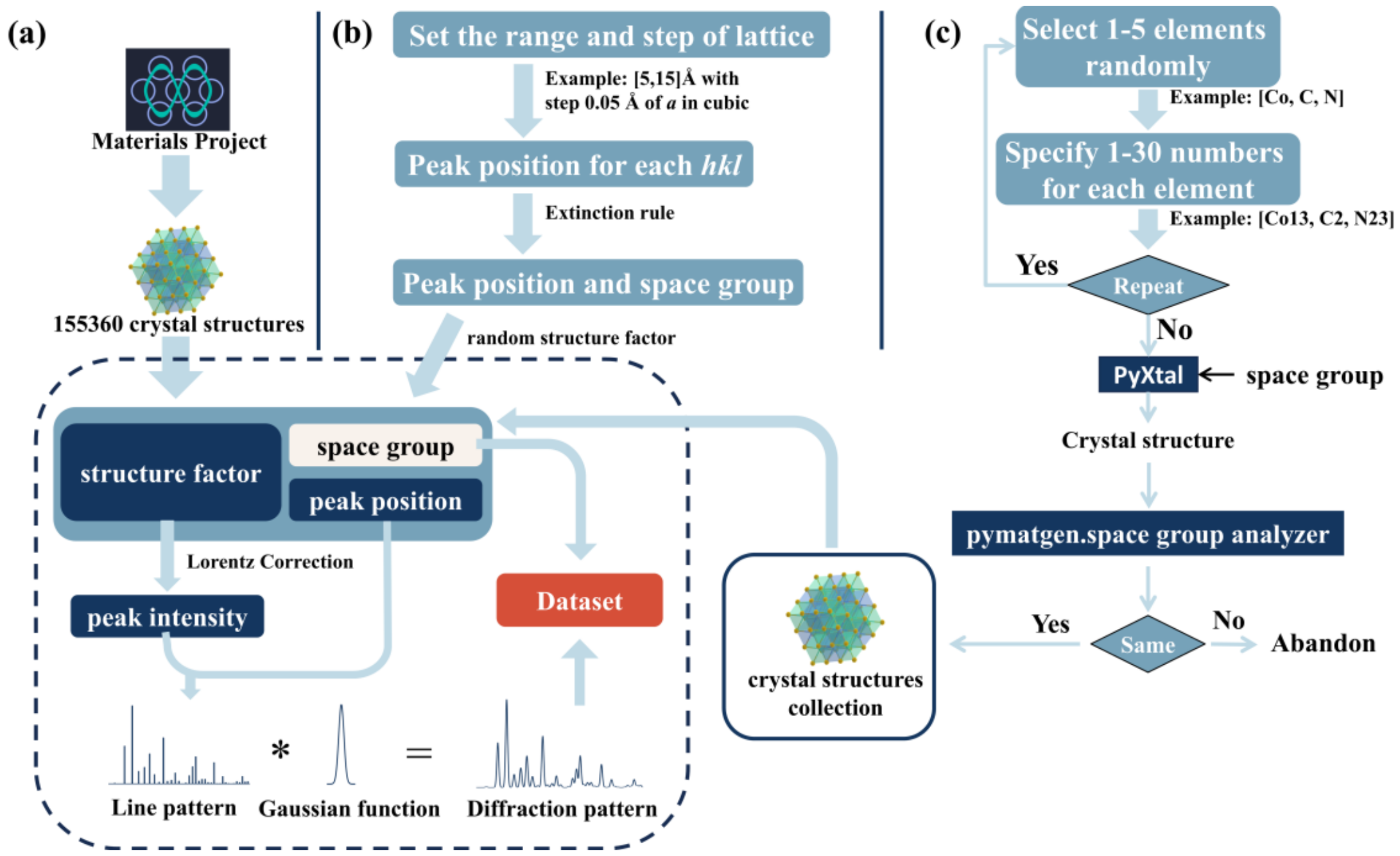

**Figure 1**. The process of generating the (a) Crystallographic Database Derived Dataset (CDDD), (b) Uniformed Lattice Based Dataset (ULBD) and (c) Generated Crystal Derived Dataset (GCDD).

# 4. the Disadvantage of Crystallographic Database Derived Dataset

## 4.1 Feature bias within the space group

In the crystallographic database of the Materials Project, there are obvious biases in lattice parameters within each space group. Two examples are shown in **Fig. 2**; each type of lattice parameter tends to cluster around one or several values rather than following a uniform or normal distribution. These biases in lattice parameters lead to biased features in the powder diffraction patterns. **Equation 1** gives the relationship between peak positions and lattice parameters, indicating that the *hkl* peak positions are determined only by the lattice parameters *a, b, c, α, β, γ*. In addition, the diffraction patterns from the structures with the same space group must obey the same extinction rules. Consequently, as shown in **Fig. 3**, the structures with similar lattice parameters in the same space group yield similar powder diffraction patterns. Conversely, diffraction patterns from structures with significantly different lattice parameters appear markedly distinct.

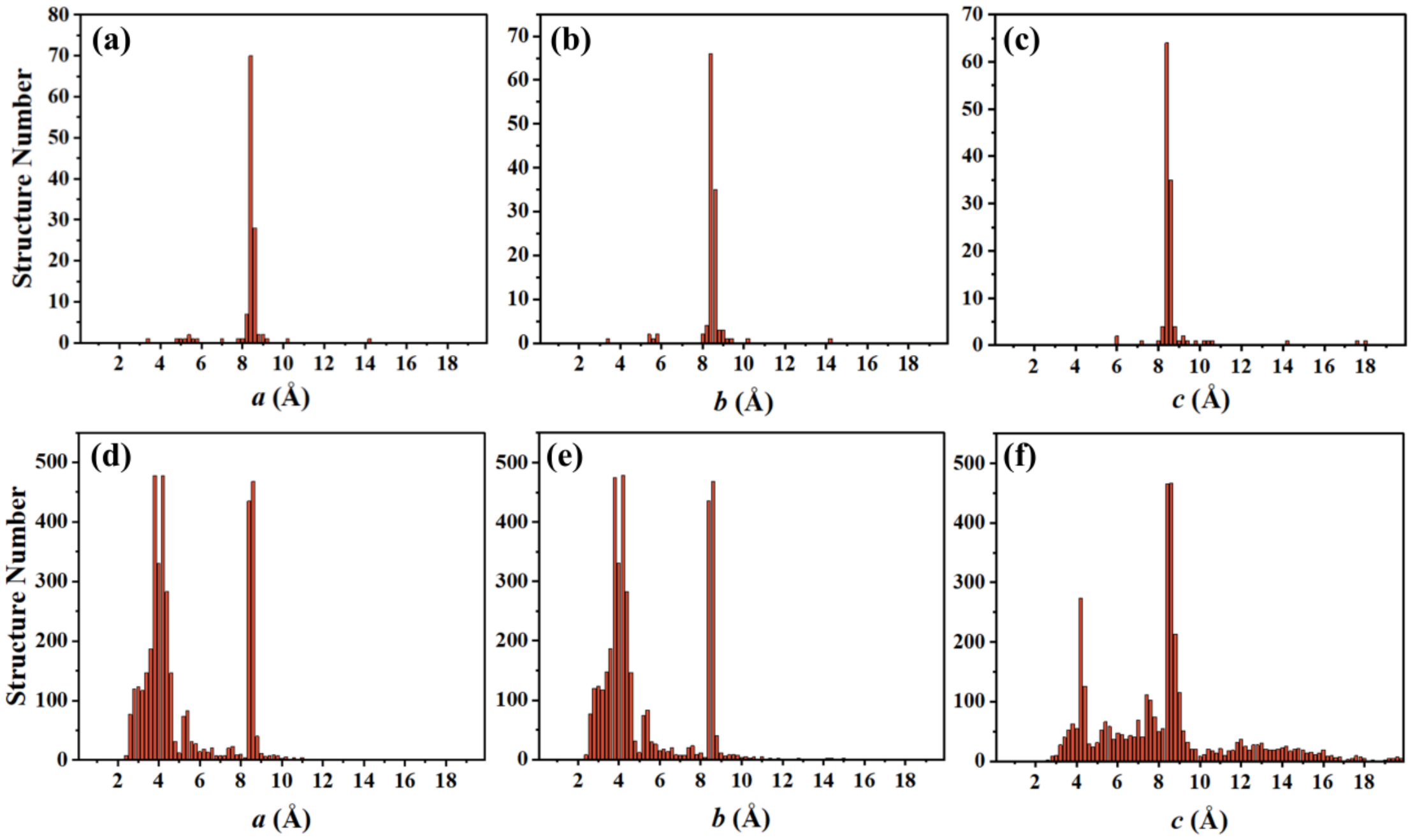

**Figure 2**. The lattice parameters' distribution histogram of the Materials Project's crystal data with space group "P 2 2 2" (a)(b)(c) and "P 4/m n c" (d)(e)(f). The bin width of the histogram is 0.2 Å. The Figure shows the parameters without α, β, γ because all of them are 90 degrees in the orthorhombic symmetry.

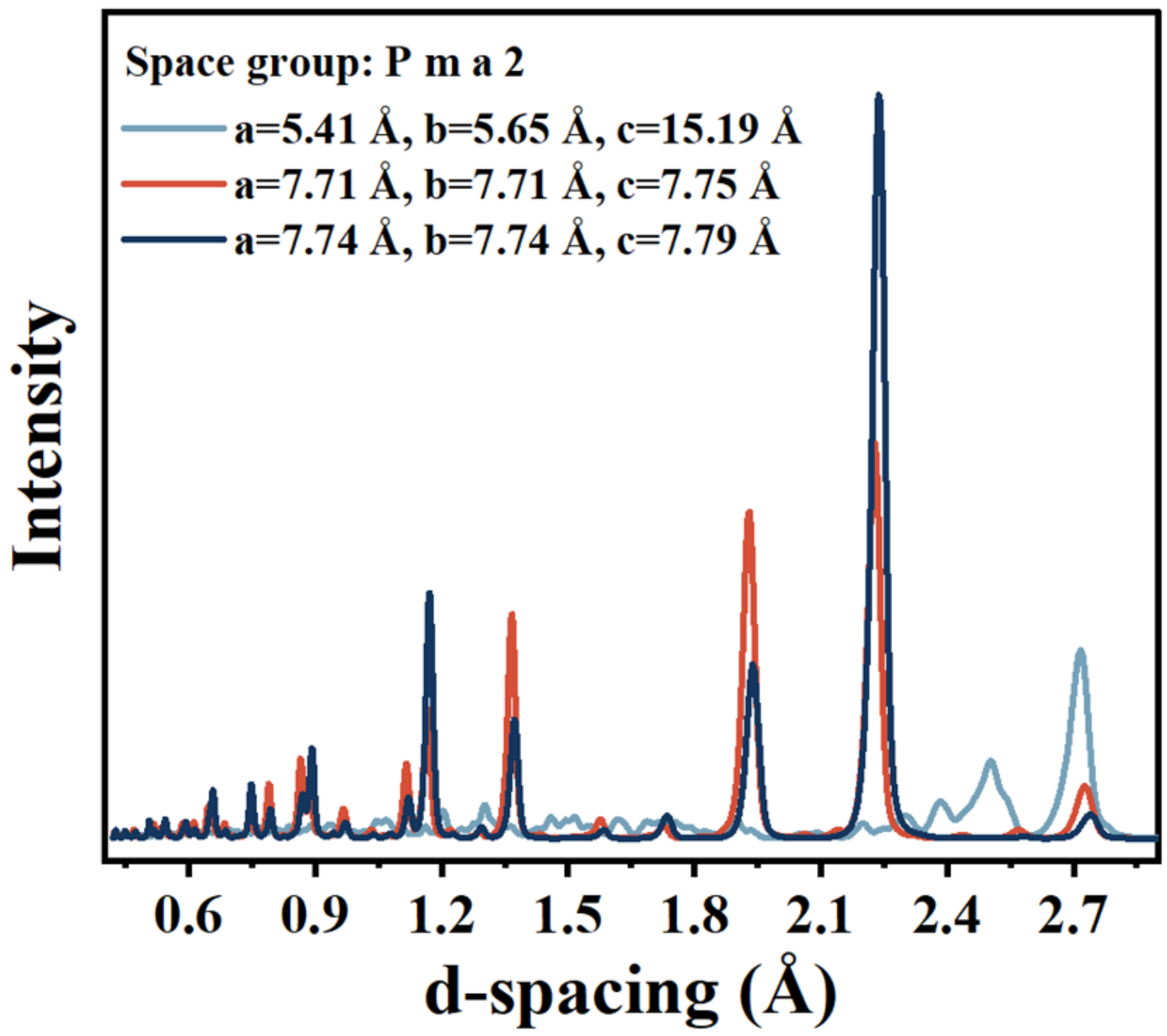

**Figure 3**. The powder diffraction patterns were calculated from the crystal with the same space group "Pma2" but with different lattice parameters in the Materials Project. The labels only show the *a, b, c* because this space group belongs to the orthorhombic, in which α=β=γ=90º.

The direct consequence of such inherent feature bias is a weak generalization ability because the model cannot infer the correct symmetry laws from biased samples. Here, CDDD and GCDD were used to assess their influence. Compared to CDDD, the GCDD features an asymmetric Gaussian-like distribution of lattice parameters in each space group, as shown in **Fig. 4**. The CDDD was split into a training set and a test set with a 9.5:1 ratio for each space group. A CNN model was trained using the CDDD training set. It was then tested on both the CDDD test set and the GCDD test set. The resulting confusion matrices are shown in **Fig. 5(a)(b)**. An obvious performance decrease can be observed on the GCDD, with accuracy much lower than that on CDDD, as shown in **Fig. 5(c)**. This indicates the weak generalization ability of the model trained on CDDD.

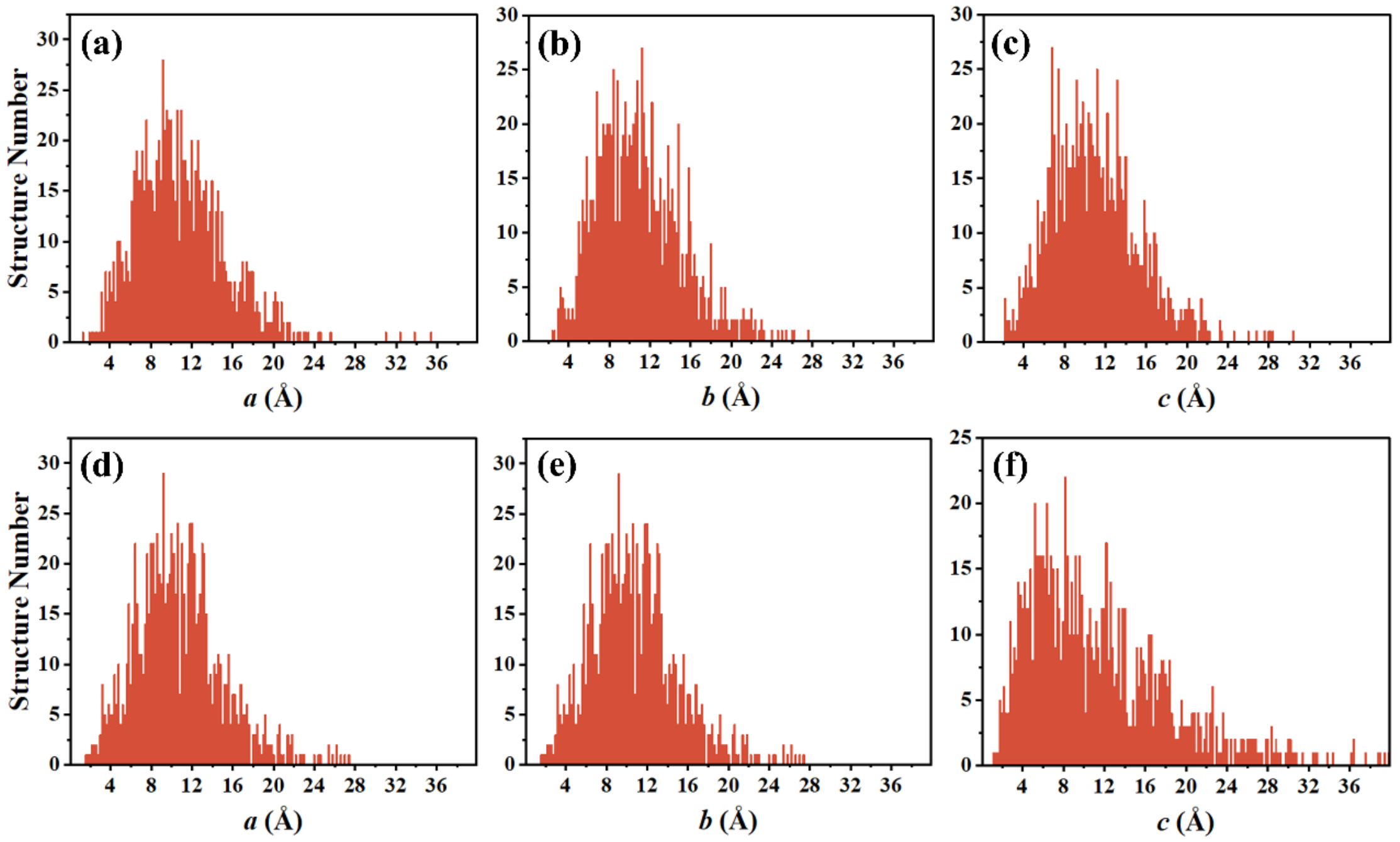

**Figure 4**. The lattice parameters' distribution histogram of the GCDD with space group "P 2 2 2"(a)(b)(c) and "P 4/m n c"(d)(e)(f). The bin width of the histogram is 0.2 Å. Compared to Fig. 2, the distribution is closer to an unsymmetric Gaussian distribution.

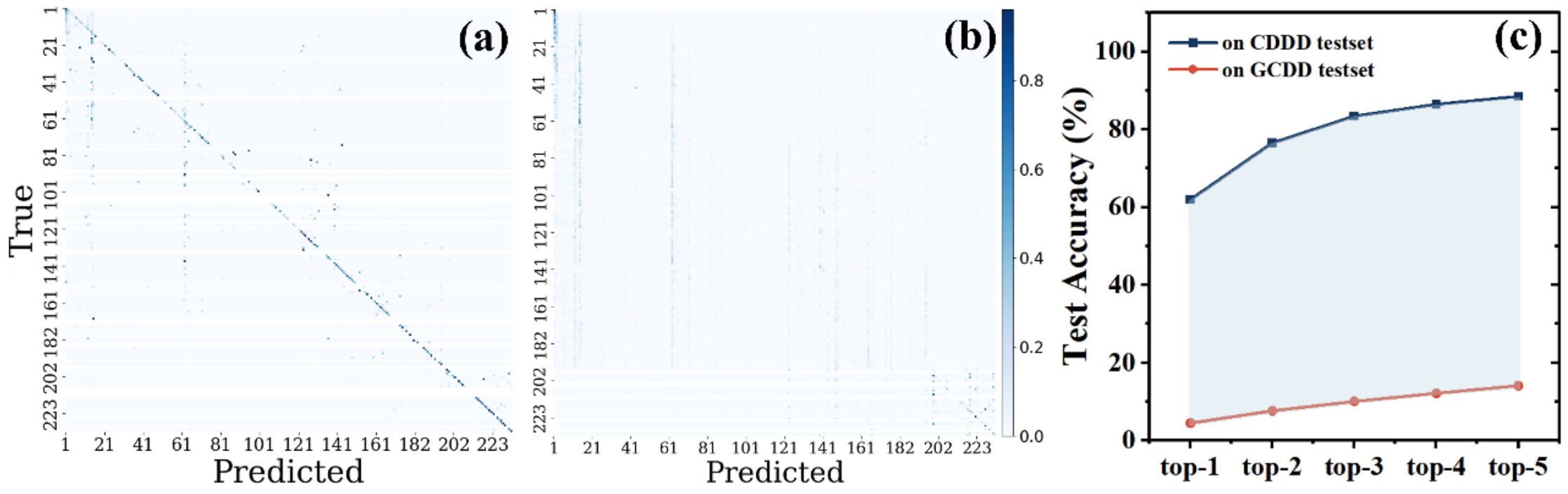

**Figure 5**. Confusion matrix illustrating the predictions for the CDDD test set (a) and the GCDD test set (b) made by the CNN model trained on the CDDD training set. (c) The top-1 to top-5 accuracy on the CDDD test set (blue point line) and GCDD test set (red point line).

## 4.2 Imbalance in the Quantity of Structures Across Space Groups

An imbalance in the number of samples across space groups is evident in the crystallographic database, as shown in **Fig. 6**. Some space groups contain up to 10,000 structures, while others contain none. This imbalance in quantity could adversely affect the model's performance.

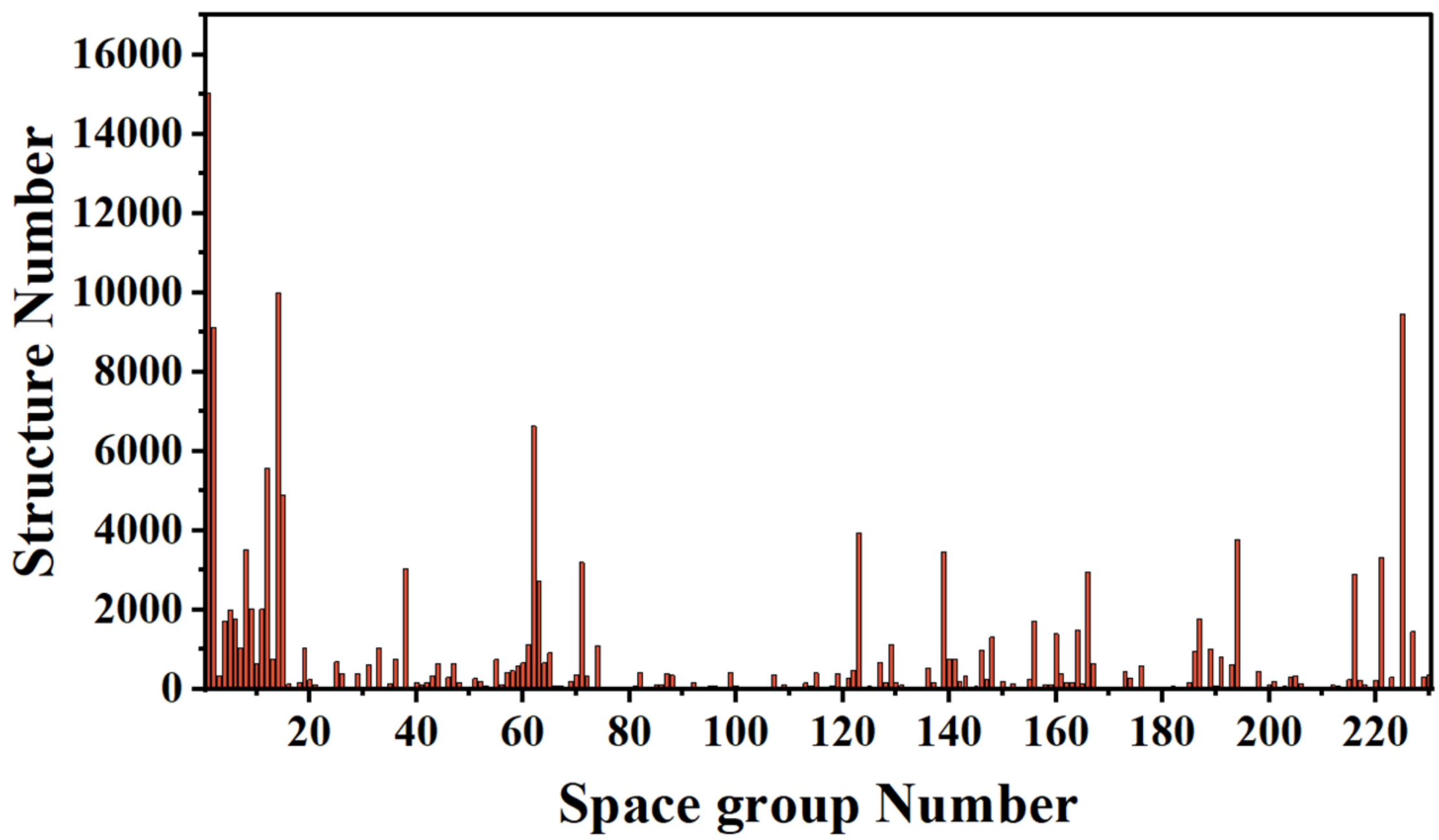

**Figure 6**. The structure amount distribution between space groups in the crystallographic database of Materials Project. The space group 168 and 207 don't have any structure.

One detrimental effect is that the model tends to favor space groups with more samples in the training set. This bias is evident in the vertical lines seen in the confusion matrix in **Fig. 5(a)(b)**. **Fig. 7(a)** displays the confusion matrix from **Fig. 5(a)** again and compares it with the distribution of structures across space groups in the CDDD. The vertical lines correspond exactly to space groups with a higher number of structures, clearly demonstrating the model's prediction bias. As a comparison, the GCDD was also split into a training set and a testing set with a ratio of 9.5:1 for each space group. A CNN model with the same architecture was trained using this training set and tested on the testing set. The resulting confusion matrix is shown in **Fig. 7(b)**, along with the distribution of structures among space groups in the GCDD training set. The absence of vertical lines indicates that the distribution in the GCDD is much more balanced than in the CDDD.

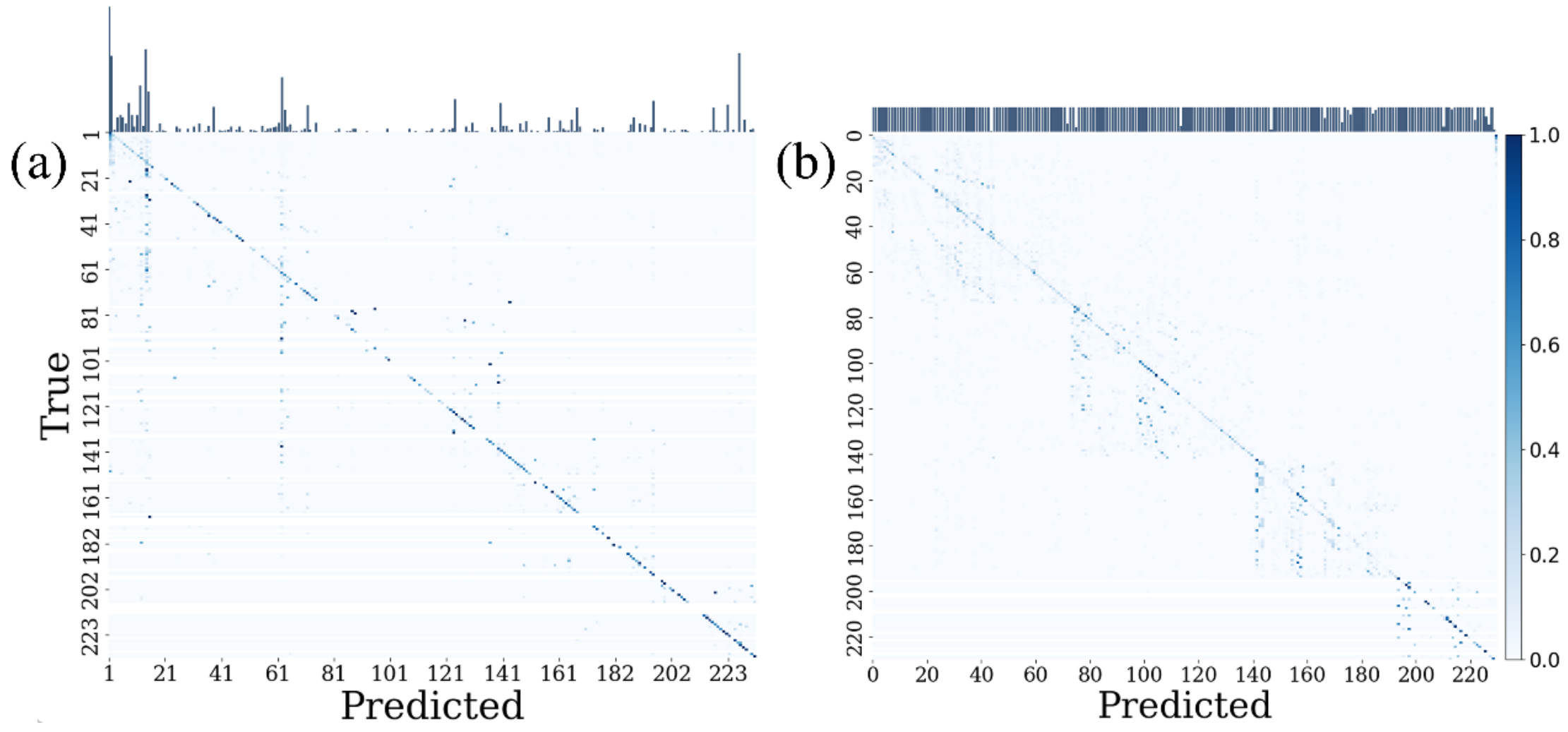

**Figure 7**. (a) Confusion matrix that illustrates the predictions for the CDDD test set made by the CNN model trained on the CDDD training set, together with the data distribution of CDDD. (b) Confusion matrix that illustrates the predictions for the GCDD test set made by the CNN model trained on the GCDD training set, together with the data distribution of GCDD.

The other detrimental effect is that a model trained on an imbalanced training set may exhibit artificially higher prediction accuracy compared to one trained on a balanced set. Here, the ULBD-cubic was designed to validate this effect. The ULBD-cubic was configured with a uniform lattice range of [5, 15] Å and a step of 0.05 Å for each space group. In contrast, the ULBD-cubic-imbalanced was designed to have each space group's lattice range match the corresponding range in the CDDD, resulting in different sample numbers across space groups. Both datasets were split into a training and a validation set with a 4:1 ratio for each space group, and two CNN models were trained on the two datasets. The training process is shown in **Fig. 8**. **Fig. 8(a)** shows that the model trained on ULBD-cubic achieved an accuracy of about 47.5% and appeared overfitting around the $50^{th}$ epoch. Meanwhile, **Fig. 8(b)** shows that the model trained on ULBD-cubic-unbalanced reached an accuracy of about 60% at first and then improved after the $20^{th}$ epoch, and finally up to over 80%. Note that the only difference between ULBD-cubic and ULBD-cubic-unbalanced is the imbalance in data quantity among different space groups. This result demonstrates that an imbalanced training set can give an unreliable higher prediction accuracy.

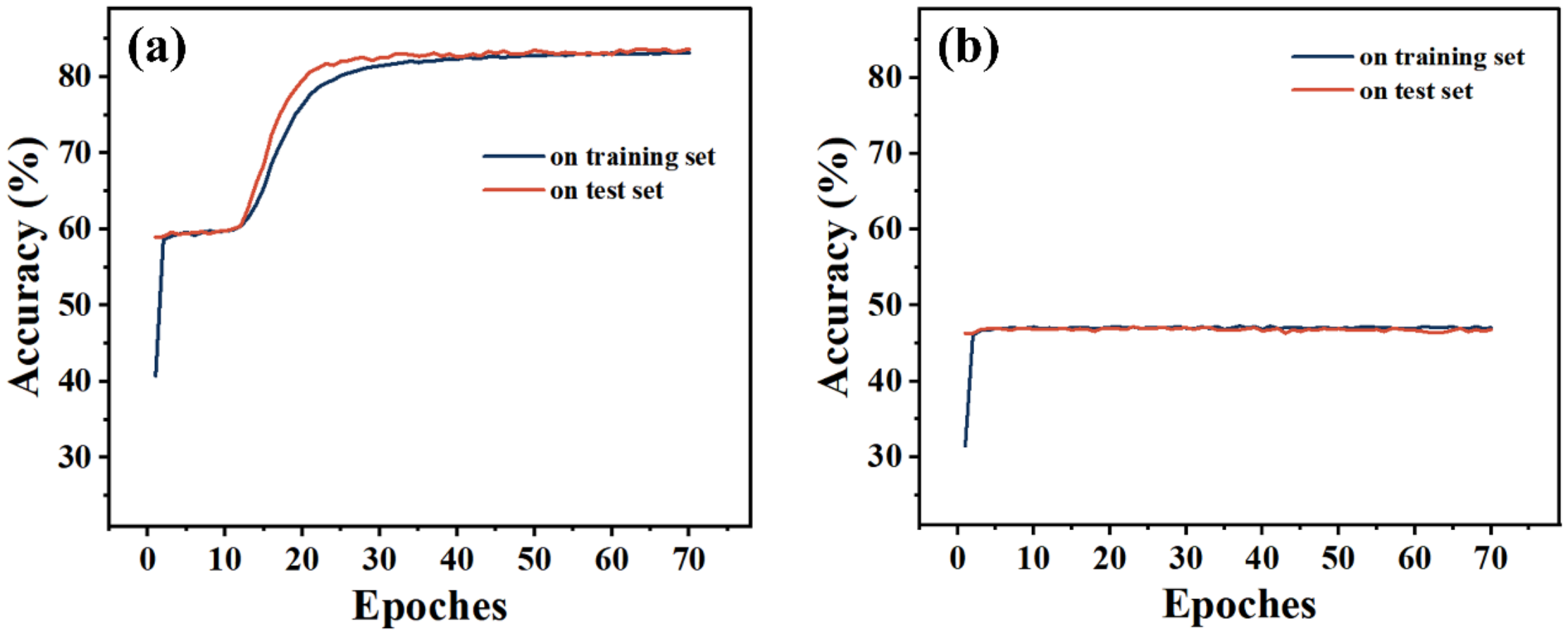

**Figure 8**. (a) The changing of top-1 accuracy during the training process on the ULBD-cubic-imbalanced. (b) The changing of top-1 accuracy during the training process on the ULBD-cubic.

## 5. Reasonable space group prediction in consistent with extinction laws

To mitigate the negative effects observed with CDDD, the ULBD, which features uniform space group and lattice distribution, should be adopted. In this section, a CNN model was trained on the ULBD-cubic, ULBD-tetragonal, and ULBD-tri/hexagonal separately for validation. In ULBD, peak intensities are random, and extinction rules determine the space group. The theoretical maximum prediction accuracy can be calculated according to the following considerations: space groups sharing the same extinction rule can't be distinguished, while those with unique extinction rules can be distinguished completely. **Tab. S1, S2, S3** organize space groups in cubic, tetragonal, and trigonal+hexagonal symmetries by extinction rules. **Tab. I** details the theoretical maximum prediction accuracy for cubic, tetragonal, and trigonal+hexagonal symmetries, from top-1 to top-5 predictions. The ULBD-cubic, ULBD-tetragonal, and ULBD-tri/hexagonal datasets have been split into training set and test set with a 5:1 ratio for each space group. Then the test set was used to evaluate the trained model's prediction accuracy for space groups from top-1 to top-5 predictions. The results, shown in **Fig. 9**, reveal an impressive outcome: all prediction accuracies from top-1 to top-5 closely match the theoretical maximum prediction accuracy. Additional tests examined how accuracy scales with the training data set size. The size of ULBD is controlled by the step of lattice changing and the number of random patterns per lattice. As shown in **Fig. S1**, when the lattice step size is less than or equal to 0.1 Å and the number of patterns per lattice is greater than or equal to 5, the model achieves the theoretical maximum accuracy.

**Table I**. The theoretical top prediction accuracy of space group in cubic, tetragonal, and trigonal+hexagonal from top-1 to top-5.

| | Top-1 | Top-2 | Top-3 | Top-4 | Top-5 |
|---|---|---|---|---|---|
| Cubic | 47.2% | 72.2% | 80.5% | 88.9% | 97.2% |
| Tetragonal | 45.6% | 70.6% | 83.8% | 88.2% | 91.2% |
| Trigonal+Hexagonal | 17.3% | 34.6% | 48.1% | 59.6% | 69.2% |

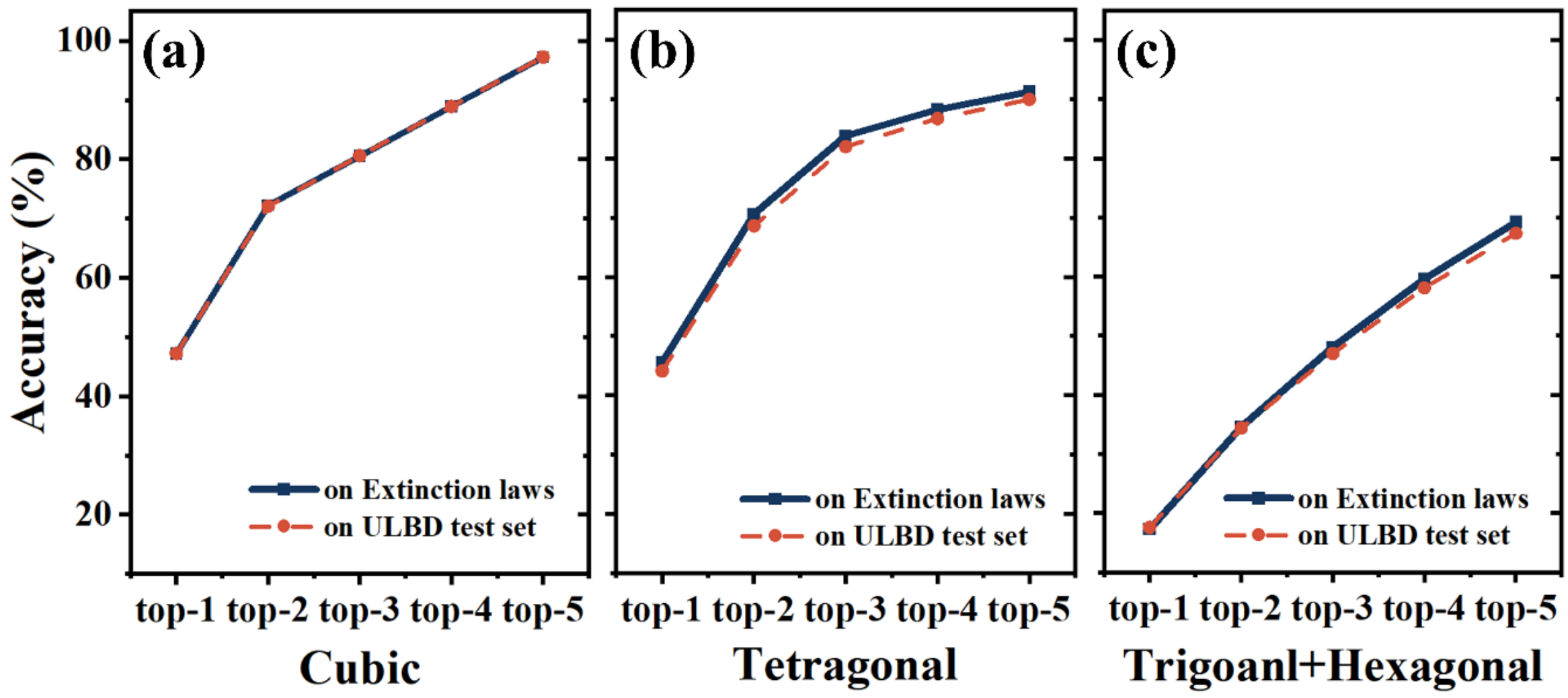

**Figure 9**. The top-1 to top-5 space group prediction accuracy of CNN model in (a) cubic, (b) tetragonal, and (c) trigonal+hexagonal symmetries. The theoretical top prediction accuracies are also illustrated for comparison.

The model's generalization capability was evaluated within the cubic, tetragonal, and trigonal+hexagonal symmetries. It is important to note that we reorganized the space groups into a new classification for this evaluation, where space groups sharing the same extinction rule are relabeled as the same class. In this classification, the theoretical highest top-1 accuracy is 100%. The evaluation category is shown in **Fig.10(a)**. Both the ULBD and CDDD datasets were split into training and test sets. Two models, named ULBD Model and CDDD Model, were trained using their respective training sets. Both the ULBD test set and CDDD test set are used to test the accuracy of the two models. In the ULBD, all peak intensities for patterns with identical lattices are generated randomly, ensuring that each pattern is unique and distinct from those in the CDDD. The testing results, shown in **Fig.10(b)(c)(d)**, demonstrate that the ULBD Model exhibits excellent generalization capability, with almost no accuracy decrease on the CDDD test set. In contrast, the CDDD model shows a significant drop in accuracy on the ULBD test set. The result indicates that the model trained on ULBD has superior generalization capability compared to the one trained on CDDD.

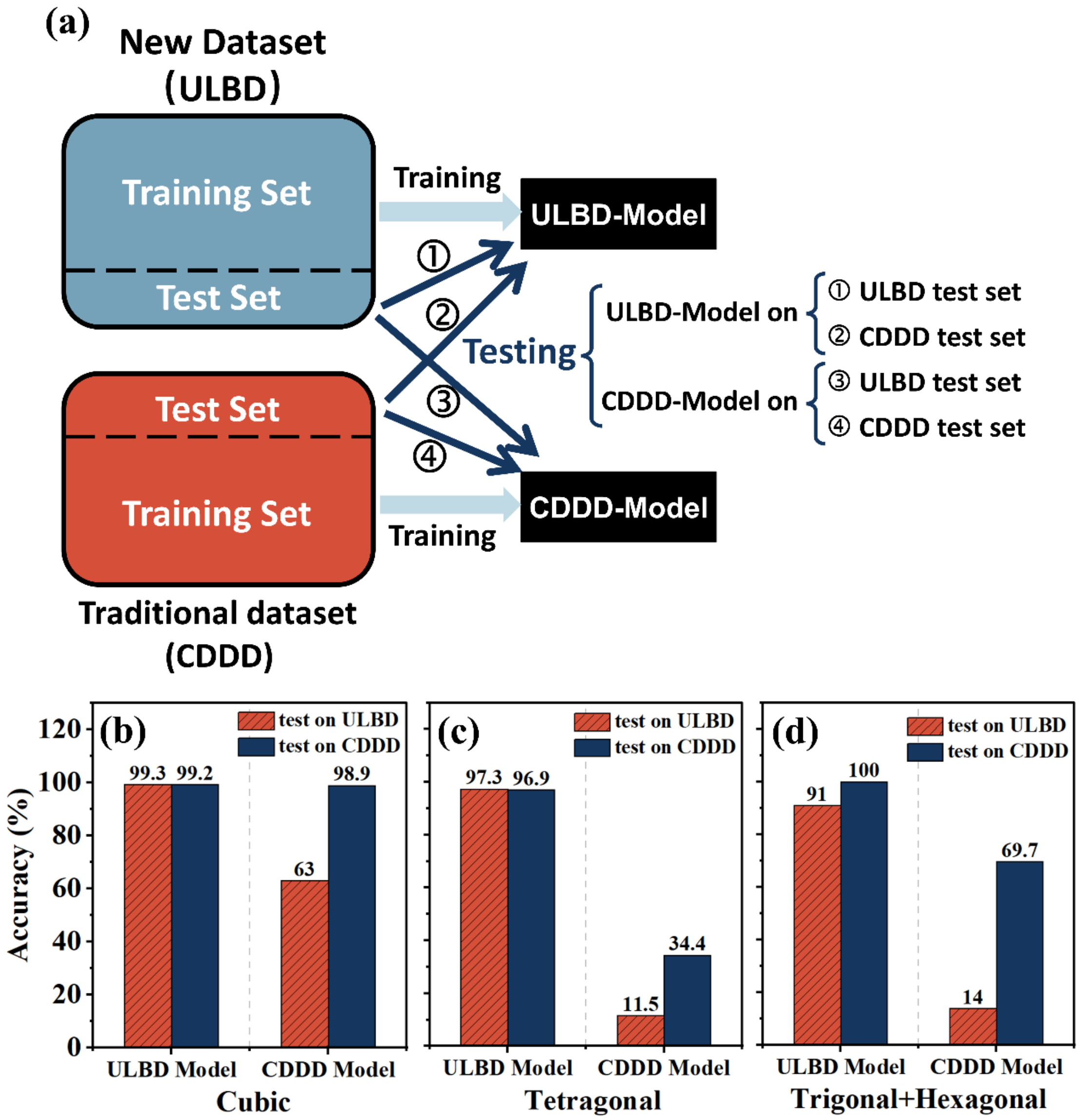

**Figure 10**. (a) The category of testing the generalization capability of the model trained by ULBD (ULBD Model). The same model but trained by CDDD (CDDD Model) also was tested for comparison. (b) The accuracy of space group prediction in cubic symmetry. Both two models were tested on ULBD test set (red) and CDDD test set (blue) separately. The ULBD Model keeps almost same prediction accuracy on the two datasets, but the CDDD Model presents a obvious decrease on the ULBD. The same result also can be observed on the prediction in (c) tetragonal symmetry and (d) trigonal+hexagonal symmetry.

## 6. Conclusion and outlook

In this paper, we demonstrate the limitations of using crystallographic databases as datasets for training deep learning models. Furthermore, we propose a novel dataset design category and a model performance evaluation method grounded in crystallographic laws. This study represents a breakthrough in AI-based space group prediction by prioritizing dataset design over model architecture or training strategies. This underscores the critical role of well-designed datasets in AI applications for scientific data analysis. However, several challenges remain to be addressed. For instance, the applicability of our approach to low-symmetry space group prediction requires further validation. In addition, as symmetry decreases, the dataset size increases exponentially, demanding greater computational resources. In future work, we will validate our strategy for low-symmetry predictions and apply our findings to real experimental data analysis.

The repository can be accessed at the following URL: https://github.com/unitsec/Getting_Dataset_for_Space_Group_Prediction.git. This repository includes code for generating ULBD, CDDD, and GCDD, along with test scripts for evaluating the performance of the ULBD model.

**Acknowledgments**
This work was supported by the National Key R&D Program of China (Grant No. 2023YFA1610000), and the National Natural Science Foundation of China (Grant No. 12305342).